\relax
\documentclass[letterpaper]{article} 
\usepackage{ijcai22}  
\usepackage{times}  
\usepackage{helvet} 
\usepackage{courier}  
\usepackage[hyphens]{url}  
\usepackage{graphicx} 
\usepackage{natbib}  
\usepackage{caption} 
\usepackage[utf8]{inputenc} 
\usepackage[T1]{fontenc}    
\usepackage{url}            
\usepackage{booktabs}       
\usepackage{amsfonts}       
\usepackage{nicefrac}       
\usepackage{microtype}      

\usepackage{amsthm}
\usepackage{enumitem}

\usepackage{url}
\usepackage[utf8]{inputenc}
\usepackage{graphicx}
\usepackage{amsmath}
\usepackage{booktabs}
\usepackage{algorithmicx}
\usepackage{algorithm}
\usepackage{amssymb}
\usepackage{multicol}
\usepackage[font=small]{subcaption}
\usepackage{enumerate}
\usepackage{enumitem}
\usepackage[noend]{algpseudocode}
\usepackage{colortbl}
\usepackage[switch]{lineno} 
\usepackage{bm}

\usepackage[cal=boondox]{mathalfa}
\usepackage{mathrsfs}

\newtheorem{theorem}{Theorem}

\newtheorem{definition}[theorem]{Definition}

\newtheorem{remark}[theorem]{Remark}
\theoremstyle{definition}
\newtheorem{example}{Example}

\newcommand{\eat}[1]{}

\title{Summary Markov Models for Event Sequences}

\author {
    Debarun Bhattacharjya$^1${\normalfont,} 
  Saurabh Sihag$^2${\normalfont,} 
  Oktie Hassanzadeh$^1${\normalfont,} 
  Liza Bialik$^3$
  \affiliations
    $^1$ IBM Research \footnote{
    Distribution Statement ``A" (Approved for Public Release, Distribution Unlimited)}, 
    $^2$ University of Pennsylvania, $^3$ University of Massachusetts, Amherst \\
    \emails
    \{debarunb,hassanzadeh\}@us.ibm.com, sihags@pennmedicine.upenn.edu,
    rbialik@cs.umass.edu
}

\begin{document}

\maketitle

\begin{abstract}
Datasets involving sequences of different types of events without meaningful time stamps are prevalent in many applications, for instance when extracted from textual corpora. We propose a family of models for such event sequences -- summary Markov models -- where the probability of observing an event type  depends only on a summary of historical occurrences of its influencing set of event types. This Markov model family is motivated by Granger causal models for time series, with the important distinction that only one event can occur in a position in an event sequence. We show that a unique minimal influencing set exists for any set of event types of interest and choice of summary function, formulate two novel models from the general family that represent specific sequence dynamics, and propose a greedy search algorithm for learning them from event sequence data. We conduct an experimental investigation comparing the proposed models with relevant baselines, and illustrate their knowledge acquisition and discovery capabilities through case studies involving sequences from text.
\end{abstract}

\section{Introduction}\label{sec:intro}

Numerous applications require interpretable models for capturing dynamics in \emph{multivariate event sequences}, i.e. sequences of different types of events \emph{without time stamps}. 
The classic $k^{th}$-order Markov chain captures such dynamics, where the probability of observing a particular event type 
depends on the preceding $k$ positions. 
Choosing a large $k$ could however result in a blowup of the state space and over-fitting,
while a small $k$ ignores potentially important older events. 
In this paper, we 
wish to learn
the specific influencers of a particular event type of interest in an event sequence, i.e. the types of events that most affect its probability of occurrence.
We are particularly interested in learning such potential influencers for knowledge discovery in the \emph{low-data} or \emph{noisy data} regimes.

Consider the following illustrative narrative, which is a sequence of events involving a common protagonist~\citep{chambers2008unsupervised}: [`person visits bank', `person visits restaurant', `person makes phone call', `person eats meal'].
Given enough data, understanding the effect of prior events may reveal that `person eats meal' is more likely to occur if `person visits restaurant' has occurred previously, and 
does not depend on `person visits bank' or `person makes phone call'.
As another example, consider a patient with numerous treatments who recovers from a chronic condition partially due to a much older therapy event. 
In both the examples, it is desirable to discover \emph{which} event types affect observations of interest, where the impacts may be from older occurrences. 

In this paper, we formalize a general family of models for multivariate event sequences -- \emph{summary Markov models} (SuMMs) -- 
where the probability of occurrence of a select set of event types at any position 
depends on a summary of historical occurrences of a subset of all the event types. 
Our intent is to learn this subset as well as the quantitative nature of the influence,  with the help of parent search techniques that are popular in graphical models such as Bayesian networks~\citep{pearl1988}.
SuMMs identify the impact of prior (possibly older) occurrences of only the key event types, i.e. those that determine the probability of observing any particular event type (or subset of types) of interest at any sequence position -- we refer to these as the \emph{influencing set}.
In the aforementioned examples, SuMMs should ideally be able to identify that `person visits restaurant' is an influencer for `person eats meal' in the daily schedule narrative example, and that `therapy' is an influencer for `recovery' in the healthcare example.
The emphasis here is primarily on providing interpretable insights about the dynamics of events of interest, by inferring  statistical/causal relations between event types from event sequences.

SuMMs generalize many well known Markov models, 
such as
$k^{th}$-order and variable order Markov chains.
However, since they identify a potentially much smaller influencing set, they are able to manage the state space size while accounting for information from the past. In that regard, they also generalize models for prediction in sequences that are typically regression-based and potentially enforce sparsity on event types. 
To complement prior literature, we propose two specific models within the SuMM family that use novel mappings to summarize historical occurrences for event sequences.

SuMMs are applicable when events have a  temporal order but no meaningful time stamps, e.g. they are suitable when:
1) it is natural to model an event sequence in discrete time -- such as top-story news events in a regular news feed; or 2) time-stamps for events are either too noisy or irrelevant -- such as for events recorded by an unreliable operator/machine; or 3) when time stamps are unavailable or difficult to obtain -- such as events extracted from text through NLP techniques.

\paragraph{Contributions.} In this paper, we:
1) formalize a general Markov model family for event sequences 
where a set of event types are affected by an influencing set, 
proving that a unique minimal influencing set exists for general event sequence dynamics; 2) formulate two instances from the family that are suitable for many real-world sequences; 3) propose a greedy algorithm for learning the proposed models including the minimal influencing set; 4) conduct experiments 
around probabilistic predictions for specific event types of interest, comparing with some relevant baselines; and 5) investigate knowledge discovery benefits of being able to identify influencers of individual event types through two case studies.

\section{Related Work}




\paragraph{Event Sequences in Data Mining.}
There is a large body of related work on mining patterns in sequences, with a wide range of applications related to explanation or prediction of events~\citep{mannila1997, weiss1998, rudin2012, letham2013}, recommendation systems~\citep{quadrana2018}, specification mining
~\citep{lemieux2015} and declarative process mining in information systems~\citep{diciccio2018}. Much of the work in data mining related areas has focused on the problem of efficiently discovering sub-sequences that occur frequently in the input data and meet certain criteria related to the application.
Our proposed family of models is related to finding \emph{episodes}, i.e. frequently occurring patterns, but goes further by introducing the influencing set notion and performing a graph search that leverages summary statistics from the frequent patterns.






\paragraph{Markov \& Related Models.}
These form a broad class of statistical models for prediction algorithms for discrete sequences. 
Variable order Markov models extend  concepts of Markov chains by incorporating dynamic dependence on history by  
incorporating both high- and small-order dependencies in the data~\citep{begleiter2004}. Algorithms for learning variable order Markov models borrow tools from data compression algorithms such as context tree weighting~\citep{willems1995}. 
A related class of models is that of hidden Markov models (HMM), which capture latent complexities of sequential datasets~\citep{rabiner1989} and are a special case of dynamic Bayes nets for modeling   discrete-time temporal data~\citep{dean1989,ghahramani1998,murphy2002}.


\paragraph{Other Graphical Models.}
Other related work includes discrete-time Granger time series~\citep{granger1969,eichler1999} and continuous-time graphical event models for multivariate event streams~\citep{didelez2008,gunawardana2016,bsg2018}, which also include event time stamps.
We refer to the latter data type as event streams rather than event sequences; note that the time stamps are crucial in this literature as they enable a temporal point process representation~\citep{aalen2008}.
Granger causal models for time series data typically consider continuous-valued measurements and therefore involve regression.
A crucial distinction between SuMMs and Granger-causal models is that only one event can occur in a position in a sequence, which affects the dynamics.
Chain event graphs are another related representation that model discrete processes exhibiting strong asymmetric dependence structures~\citep{collazo2018}.

\section{Model Formulation}

\subsection{
Preliminaries}

An \textbf{event sequence dataset} involves 
sequences of events of different types. Formally, 
$\mathbf{D}$ is a multiset $\{ \mathbf{D_k} \}_{k=1}^{K}$, where
$ \mathbf{D_k} = [  l_i  ]_{i=1}^{N_k}$ and event label (or type) $l_i$ at  index $i$ in the sequence is from a known label set (or alphabet), 
$l_i \in \mathcal{L}$, such that $|\mathcal{L}| = M$.
There are $K$ sequences of events in the dataset with $N = \sum_{k=1}^K {N_k }$ events. 
We are interested in how historical occurrences of some event types impact others. 

\begin{definition}
The \textbf{history} at position $i$ in an event sequence is $h_i = \{ (j, l_j) \}_{j=1}^{i-1}$.
The \textbf{history restricted to label set $\mathbf{Z} \subset \mathcal{L} $} at position $i$ only includes prior occurrences of labels from $\mathbf{Z}$; it is denoted $h^\mathbf{Z}_i = \{ (j, l_j ): j < i, l_j \in \mathbf{Z} \}$. We remove subscript $i$ when referring to a generic position.
\end{definition}


\begin{example}
\label{ex:history-def}
For event sequence $[A, A, C, B]$, history 
$h_4 = \{(1,A), (2,A), (3,C)\}$. When
restricted to 
$\mathbf{Z} = \{A, B \}$, 
$h^\mathbf{Z}_4 = \{(1,A), (2,A)\}$. ($B$ at position $4$ is excluded.) 
\end{example}
Note that we explicitly retain indices of relevant labels in history for modeling flexibility. 
For instance, one may wish to ignore older prior events and only consider the
most recent $k$ positions, or conversely to ignore the most recent positions so as to model delay in the dynamics.


\begin{definition}
A sequence \textbf{summary function} $s(\cdot)$ for label set $\mathbf{Z}$ maps any restricted history $h^\mathbf{Z}$ at any sequence position to some \textbf{summary state} $s_\mathbf{Z}$ in a discrete range $\Sigma_\mathbf{Z}$. 
History $h$ is  \textbf{consistent} with state $s_\mathbf{Z}$ if the summary function applied to $h$ restricted to $\mathbf{Z}$ results in $s_\mathbf{Z}$, i.e. $s(h^\mathbf{Z}) = s_\mathbf{Z}$.
\end{definition}

A summary function is intended to summarize any possible history in a sequence into a (relatively) smaller number of states, which
enables learning from limited data. The following example illustrates the ability to determine consistency between summary states for different label sets.

\begin{example}
\label{ex:consistency}
Consider a summary function $s(\cdot)$ that results in binary instantiations over a label set that specify whether a label occurred at least once in the restricted history. (In Section~\ref{sec:summ}, we consider this function for a proposed model). 
For $\mathbf{Z} = \{ A,B,C\}$ and
history $h_4$ from Example~\ref{ex:history-def},
the summary state is $\{a, \bar{b}, c \}$, 
which specifies that $A$ and $C$ occur in history but $B$ does not. This history is also consistent with summary $\{a, c \}$ over label set $\{A,C\}$ 
but not with $\{a, b\}$ over $\{A,B\}$.
\end{example}


\subsection{Event Sequence Dynamics}

A sequential process over labels in $\mathcal{L}$ where the global dynamics in the multivariate event sequence are conditionally homogeneous given the history can be captured by a parameterization using probabilities
$\Theta = \{ \Theta_X:  X \in \mathcal{L} \}$, 
$ \Theta_X = \{ \theta_{x|h}\}$ 
s.t. $\sum_{X \in \mathcal{L}} {\theta_{x|h}} = 1$ for all possible histories $h$, where $\theta_{x|h}$ is the probability of event label $X$ occurring at any position in the sequence given history $h$.
While considering
the dynamics of a subset of the event labels $\mathbf{X} \subseteq \mathcal{L}$, 
we introduce a corresponding random variable  denoted $\mathcal{X}$ which has a state for each label in $\mathbf{X}$ and a single state for when the label belongs to $\mathcal{L}\setminus \mathbf{X}$, if the set $\mathcal{L}\setminus \mathbf{X}$ is not empty. 
Thus, when $\mathbf{X}$ is a strict subset of all labels ($\mathbf{X} \subset \mathcal{L}$), there are $|\mathbf{X}| + 1$ states of $\mathcal{X}$; we denote these as $\mathcal{x}$. 
We use $\tilde{\Theta}_\mathbf{X}$ to denote the sum of probabilities over label set $\mathbf{X} \subseteq \mathcal{L}$, i.e.
$\tilde{\Theta}_\mathbf{X} =
\{ \tilde{\theta}_{x|h}\}$ where 
$ \tilde{\theta}_{x|h} = \sum_{X \in \mathbf{X}} { \theta_{x|h}}$.




\begin{definition}
Label sets $\mathbf{U}$ and $\mathbf{V} = \mathcal{L} \setminus \mathbf{U}$ are \textbf{influencing}
and \textbf{non-influencing} sets
for event labels $\mathbf{X}$  
under summary function $s(\cdot)$ if for all $s_\mathbf{U} \in \Sigma_\mathbf{U}$, $\tilde{\theta}_{x|h} = \tilde{\theta}_{x|h'}$ for all $h,h'$ consistent with $s_\mathbf{U}$. $\mathbf{U}$ is \textbf{minimal} if the condition cannot be satisfied after removing any label in $\mathbf{U}$.
\end{definition}


As per the definition above, historical occurrences of 
non-influencers
of $\mathbf{X}$ under summary function $s(\cdot)$ do not affect the probability of observing labels from $\mathbf{X}$ at any sequence position.
(Please see 
Appendix~A\footnote{Appendices are in the arXiv version of the paper.}
for proofs.)

\begin{theorem}
\label{thm:unique}
There is a unique minimal influencing set for any set of labels $\mathbf{X} \subseteq \mathcal{L}$ 
and summary function $s(\cdot)$ 
for any conditionally homogeneous event sequence dynamics $\Theta$.
\end{theorem}




It is natural to pose the question of whether it is possible to formulate a directed (potentially cyclic) graphical representation for global dynamics in event sequences, similar to Granger causal related graphs~\citep{eichler1999,didelez2008}. 

\begin{definition}
\label{def:simplify}
An 
event sequence
parameterization $\Theta$ is said to \textbf{simplify} according to a directed graph $\mathcal{G}$ over labels $\mathcal{L}$ and summary function $s(\cdot)$ if the parents of each node $X \in \mathcal{G}$ are influencing sets for their children individually, under $s(\cdot)$.
\end{definition}

\begin{theorem}
\label{thm:simplify}
For any directed graph $\mathcal{G}$ over labels $\mathcal{L}$ and a summary function $s(\cdot)$,
there exists some event sequence parameterization $\Theta$ that simplifies according to $\mathcal{G}$ and $s(\cdot)$.
\end{theorem}

Simplification is analogous to factorization in Bayes nets~\citep{pearl1988}; it is used to formalize the situation where event sequence dynamics satisfy the parameterization constraints implied by some underlying graph.
Note that any parameterization $\Theta$ simplifies according to the fully connected directed graph including self loops (which indicate self-influence), just like any distribution over random variables factorizes according to a fully connected Bayes net. 

\subsection{Summary Markov Models (SuMMs)
}

In many applications, one is interested in modeling the dynamics of an individual label (or small set of labels) of significance to the modeler, with the intent of finding the minimal influencing set. This is of practical importance particularly for  knowledge discovery  using event sequence datasets.
Theorem~\ref{thm:unique} highlights that a unique minimal influencing set exists for any label set $\mathbf{X} \in \mathcal{L}$ for any summary function $s(\cdot)$.

We therefore propose a family of models that relates occurrences of prior event labels belonging to a subset $\mathbf{U}$ to the random variable $\mathscr{X}$ corresponding to label set $\mathbf{X} \in \mathcal{L}$. 
The idea is to enable summarizing any restricted history with respect to these influencing labels $\mathbf{U}$ for random variable $\mathcal{X}$, making it conditionally independent of other histories given the summary at any position. 
Thus,
$P(\mathcal{x}|h_i) = P(\mathcal{x}|h^\mathbf{U}_i) = P(\mathcal{x}|s_\mathbf{U})$ for any state $\mathcal{x}$ of the random variable $\mathcal{X}$ and for any position $i$ in an event sequence, where $s_\mathbf{U}$ is the summary state.
A general formulation follows:
\begin{definition}
A \textbf{summary Markov model} (SuMM) for event label set $\mathbf{X} \subseteq \mathcal{L}$ (and corresponding random variable $\mathcal{X}$) includes a summary function $s(\cdot)$, a set of influencing labels $\mathbf{U}$ and probability parameters $\theta_{\mathcal{x}|s_\mathbf{U}}$ for each state of $\mathcal{X}$ and each summary state $s_\mathbf{U} \in \Sigma_\mathbf{U}$.

\end{definition}

\eat{
\begin{figure*}[t!]
\centering
\includegraphics[width=0.85\linewidth]{fig_combined.png}
\caption{(a) Graphical representation for a SuMM, where $E_i$ is the observed event label, $h^{\mathbf{U}}_i$ the restricted history, $s^{\mathbf{U}}_i$ the summary state and $\mathcal{X}_i$ the random variable for label set $\mathbf{X}$, all at position $i$ in the sequence. Concentric ovals represent deterministic funtions.
(b) Illustrative event sequence over labels $\{A,B,C \}$ where $X = C$ is of interest. Also shown are instantiations $\mathbf{u}$ for BSuMM, $\mathbf{o}$ for OSuMM at all $C$ occurrences, for influencing set $\mathbf{U} = \{A,B\}$ and look-back of $3$ positions.
}
\label{fig:combined}
\end{figure*}
}

\begin{figure}
    \centering
    \includegraphics[width=\columnwidth]{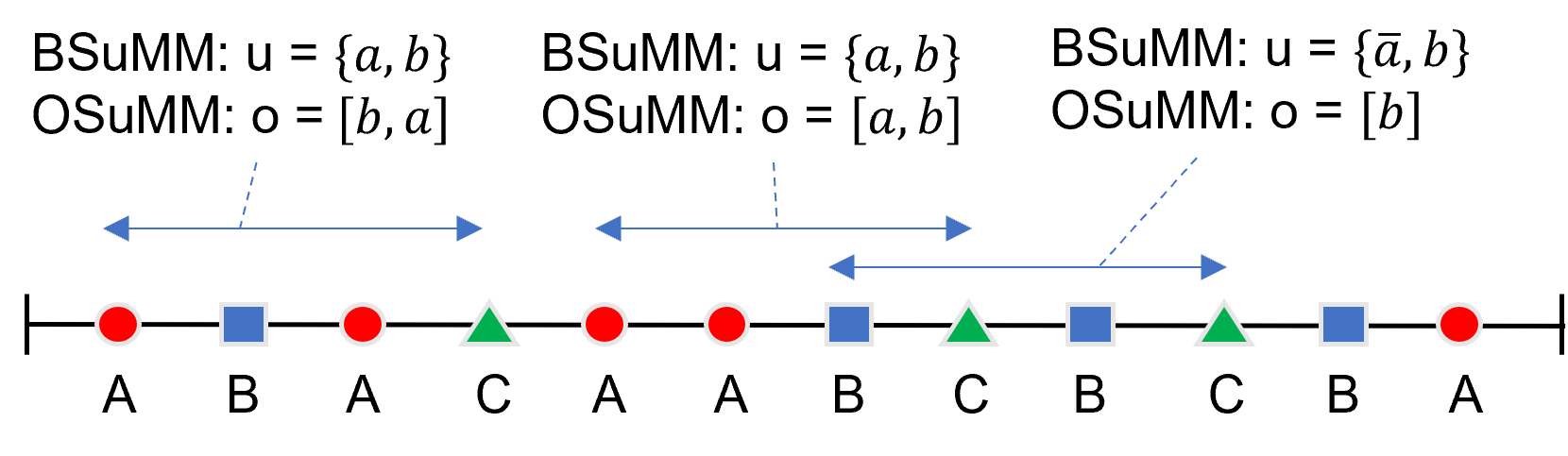}
    \caption{Illustrative event sequence over labels $\{A,B,C \}$ where $X = C$ is of interest. Also shown are instantiations $\mathbf{u}$ for BSuMM, $\mathbf{o}$ for OSuMM at all $C$ occurrences, for influencing set $\mathbf{U} = \{A,B\}$ and look-back of $3$ positions. 
    }
    \label{fig:seq_example}
\end{figure}

In practice,
if the summary function in a SuMM has range $\Sigma_{\mathbf{U}}$ that is too large, it is challenging to learn a model 
without enough instances to generalize over. This is the case for instance with $k^{th}$-order Markov chains for large $k$.

\begin{remark}
\label{rem:summ_gen}
$k^{th}$-order Markov chains and variable order Markov chain models such as context-tree weighting (CTW) are special cases of summary Markov models.
\end{remark}


SuMMs are a broad class of models for sequential dynamics around a subset of event labels, whose occurrences depend on a summary of historical occurrences of relevant event types.
The nature of the mapping from the relevant historical past $h^\mathbf{U}$ to $\Sigma_{\mathbf{U}}$ is what distinguishes various models within the broad family, which is what we expound upon next.

\subsection{Two Specific SuMMs}
\label{sec:summ}


We propose two practical models within the broader SuMMs family, suitable for studying dynamics of individual event labels in a variety of real-world datasets that involve event sequences.
In the first model,
the probability of observing a state of random variable $\mathcal{X}$ corresponding to labels $\mathbf{X}$ depends on whether one of its influencing event labels have happened or not, within some label-specific and potentially user-provided historical look-back positions. Formally:

\begin{definition}
\label{def:bsumm}
A binary summary Markov model (BSuMM) for event label set $\mathbf{X} \subseteq \mathcal{L}$ (and corresponding random variable $\mathcal{X}$) includes a set of influencing labels $\mathbf{U}$, a set of look-back positions for each influencing label $\kappa = \{ k_Z: Z \in \mathbf{U} \}$, and probabilities $\theta_{\mathcal{x}|\mathbf{u}}$ for each state of $\mathcal{X}$ and each binary vector instantiation $\mathbf{u}$ of  $\mathbf{U}$.
\end{definition}

BSuMMs are similar to Bayesian networks with binary variable parents in that there is a parameter for every state $\mathcal{x}$ and every possible parental configuration $\mathbf{u}$ from the $2^{|\mathbf{U}|}$ possibilities.
BSuMMs are simple yet suitable for a wide range of event sequence data, as we will demonstrate later. 

In a BSuMM, only the presence or absence of a parent in the relevant history has an effect, 
regardless of the order. 
For the second model, we allow for different parameters for different orders; before formalizing the model however, we introduce some notation, modifying definitions from prior work on historical orders in continuous-time models for time-stamped event streams~\citep{ogem_ijcai_20}.

\begin{definition}
\label{def:masking_fn}
A masking function $\phi(\cdot)$ takes event sequence $s = \{ (j,l_j) \}$ as input and returns a sub-sequence 
without label repetition,
$s' = \{ (k, l_k) \in s: l_k \neq l_m \text{ for } k \neq m\}$. 
\end{definition}

Since labels may recur in a sequence, a masking function $\phi(\cdot)$ reduces the number of possible orders 
to manageable levels.
In this paper, we apply a $\phi(\cdot)$ that favors more recent occurrences, consistent with 
other
Markov models. Specifically, we only retain the
last occurrence of an event label in an input sequence but we note that other choices are possible.



\begin{definition}
An order instantiation $\mathbf{o}$ for label set $\mathbf{Z}$ is a permutation of a subset of $\mathbf{Z}$. The order instantiation at index $i$ in an event sequence over $k$ preceding positions is determined by applying masking function $\phi(\cdot)$ to events restricted to labels $\mathbf{Z}$ occurring 
in positions $[\max{(i-k, 0)}, i)$. 
\end{definition}

\begin{definition}
An ordinal summary Markov model (OSuMM) for event label set $\mathbf{X} \subseteq \mathcal{L}$ (and corresponding random variable $\mathcal{X}$) and masking function $\phi (\cdot)$ includes a set of influencing labels $\mathbf{U}$, a single look-back position  $\kappa$ for all influencing labels, and probabilities $\theta_{\mathcal{x}|\mathbf{o}}$ for each state of $\mathcal{X}$ and each order instantiation $\mathbf{o}$ of  $\mathbf{U}$.
\end{definition}

\begin{example}
Figure~\ref{fig:seq_example} shows an illustrative event sequence with labels $\{ A, B, C\}$. It also highlights the instantiations $\mathbf{u}$ for BSuMM and $\mathbf{o}$ for OSuMM that would be applicable for each occurrence of label $C$ with respect to label set $\{A,B\}$ using a look-back of $3$ positions. The effect of the masking function for OSuMM can be seen at the occurrence of label $C$ at position $4$, where 
the order instantiation is $[b, a]$ because only the more recent $A$ occurrence at position $3$ is retained. For $C$'s occurrence at position $10$, the order instantiation is $[b]$ since $C$ is not a parent of itself.  
\end{example}

Although an OSuMM is more expressive than a BSuMM with the same influencing set $\mathbf{U}$, since it allows for a parameter for every order instantiation rather than a configuration relying on the presence or absence of an influencing label in the relevant history, it comes at the price of learnability for datasets of limited size; there are $\sum_{i=0}^{|\mathbf{U}|} {\frac{|\mathbf{U}|!} {i!}}$ order instantiations for set $\mathbf{U}$ in general. Therefore if $|\mathbf{U}|=3$, BSuMM and OSuMM would involve $2^3 = 8$ and $\sum_{i=0}^3 {\frac{3!} {i!}} = 16$ parameters respectively.

\section{Learning Summary Markov Models}

\begin{algorithm} [t!]
\caption{Greedy Score-based Search}
\begin{algorithmic}[1]
\Procedure{InfluencerSearch}{Label set $\mathbf{X} \subseteq \mathcal{L}$, data $\mathbf{D}$, model: BSuMM/ OSuMM, look-back(s) $\kappa$, masking f\textsuperscript{n} $\phi(\cdot)$ (for OSuMM), prior param. $\alpha$, penalty $\gamma$}  
    \State $\mathbf{Pa(X)} \leftarrow \emptyset$
    ; $S^* \leftarrow - Inf$
    \For {each label $E$ in $\mathcal{L} \setminus \mathbf{Pa(X)}$}  \Comment{Forward search}
        \State $\mathbf{Pa(X)'} \leftarrow \{\mathbf{Pa(X)} \cup E \}$
    \State
     Compute  $S(\mathbf{Pa(X)'})$ (`ComputeScore') \If{$S(\mathbf{Pa(X)'})$ > $S^*$}
     \State
        $S^* \leftarrow S(\mathbf{Pa(X)'})$; $\mathbf{Pa(X)}\leftarrow \mathbf{Pa(X)'}$
        \EndIf
    \EndFor  \label{fgraph}  
    \For {each label $E$ in $\mathbf{Pa(X)}$}  \Comment{Backward search}
        \State $\mathbf{Pa(X)'} = \{ \mathbf{Pa(X)}\setminus E \}$
        \State
       Compute $S(\mathbf{Pa(X)'})$ (`ComputeScore') 
 \If{$S(\mathbf{Pa(X)'})$ > $S^*$}
    \State
    $S^* \leftarrow S(\mathbf{pa(X)'})$; $\mathbf{Pa(X)} \leftarrow \mathbf{Pa(X)'}$
    \EndIf
    \EndFor  \label{bgraph}  
    \State Return $\mathbf{Pa(X)},
    \{ \hat{\theta}_{\mathcal{x}|\mathbf{s_{Pa(X)}}} \}$
\EndProcedure
\end{algorithmic}
\begin{algorithmic}[1]
\Procedure{ComputeScore}{Label set $\mathbf{X}$, influencers $\mathbf{U}$, data $\mathbf{D}$, model: BSuMM/ OSuMM, look-back(s) $\kappa$, masking f\textsuperscript{n} $\phi(\cdot)$ (for OSuMM),  prior param. $\alpha$, penalty $\gamma$}
\State Compute  summary statistics $N(\mathcal{x};\mathbf{s})$ and $N(\mathbf{s})$ 
\begin{itemize}
    \item For BSuMM, $N(\mathcal{x}, \mathbf{u})$ requires $\kappa$ 
    \item For OSuMM, $N(\mathcal{x}, \mathbf{o})$ requires $\kappa$ and $\phi(\cdot)$
\end{itemize}
\State Compute Bayesian parameter estimates 
using 
$\alpha$
\State
Compute log likelihood at these estimates 
and score as computed in eqs~(\ref{eqn:LL}) and~(\ref{eqn:score}) using 
$\gamma$ 

\Return $\{ \hat{\theta}_{\mathcal{x}|\mathbf{s}} \}$  and score $S(\mathbf{U})$ 
\EndProcedure
\end{algorithmic}
\label{alg:fbsearch}
\end{algorithm}



The primary purpose in learning SuMMs, differentiating it from prevalent Markov models, is to identify the influencing set of labels of interest $\mathbf{X}$ from event sequence data. 
Algorithm~\ref{alg:fbsearch} presents an approach that applies to the two specific proposed models as well as others within the family.
The `InfluencerSearch' procedure is a greedy forward and backward search, which is efficient and popular in graphical models~\citep{chickering2002optimal}; it requires a sub-procedure that can compute a model's score on a dataset when set $\mathbf{U}$ is known. 


The `ComputeScore' procedure in Algorithm~\ref{alg:fbsearch} estimates conditional probability parameters $\{ \hat{\theta}_{\mathcal{x}|\mathbf{s}} \}$ and ultimately the model score. We rely on the following straightforward log likelihood computation for random variable $\mathcal{X}$ for a SuMM on an event dataset using \emph{summary statistics}:
\begin{equation}
    LL_\mathcal{X} =
    \sum_\mathcal{x} {
    \sum_\mathbf{s} 
    { \left(
    N(\mathcal{x};\mathbf{s})
    log(\theta_{\mathcal{x}|\mathbf{s}}) \right)
    } },
    \label{eqn:LL}
\end{equation}
where $N(\mathcal{x}, \mathbf{s})$ counts the number of times in the dataset where the random variable $\mathcal{X}$ was observed to be in state $\mathcal{x}$ and the historical summary in that position was $\mathbf{s}$, 
based on look-back(s) $\kappa$ that are treated as hyper-parameters in this work. 
For simplicity, note that all equations are written for a single event sequence but they extend easily to multiple sequences. Also, the dependence of the summary $\mathbf{s}$ on the influencing set $\mathbf{U}$ is hidden throughout for the sake of notational convenience.

To avoid learning zero probabilities so that a model generalizes to a test dataset, we take a Bayesian approach for estimating probability parameters 
from equation~(\ref{eqn:LL}). For a  Dirichlet prior with parameters $\alpha_{\mathcal{x}|\mathbf{s}}$, a Bayesian estimate for probability parameters is computed from summary statistics as
$\hat{\theta}_{\mathcal{x}|\mathbf{s}} = \frac{\alpha_{\mathcal{x}|\mathbf{s}} + N(\mathcal{x};\mathbf{s})}
    {\alpha_{\mathbf{s}}+N(\mathbf{s})}$,
where $N(\mathcal{x}, \mathbf{s})$ is as described earlier,  $N(\mathbf{s}) =
\sum_{\mathbf{x}} 
{ N(\mathcal{x}, \mathbf{s}) }$ and $\alpha_{\mathbf{s}} = \sum_{\mathbf{x}} {\alpha_{\mathcal{x}|\mathbf{s}}}$.
For experiments, we use a single parameter $\alpha$ as hyper-parameter, assuming that $\alpha_{\mathcal{x}|\mathbf{s}} = \alpha$ $\forall$ $\mathcal{x}, \mathbf{s}$. 

We use Bayesian information criterion (BIC) as the score for $\mathcal{X}$ for a SuMM on a dataset: 
\begin{equation}
    Score_\mathcal{X} = LL_\mathcal{X}^* - \gamma |P| \frac{\log{(N)}}{2},
    \label{eqn:score}
\end{equation}
where $\gamma$ is a penalty on complexity, generally set at a default value of $1$ unless otherwise specified, $|P|$ is the number of free model parameters and $LL_\mathcal{X}^*$ is the log likelihood from equation~(\ref{eqn:LL}), computed at the probability parameter estimates. 
Recall that $N$ is the total number of events in the dataset. The BIC score penalizes models that are overly complex.

\begin{theorem}
\label{thm:algo}
If the dynamics of 
$\mathbf{X} \in \mathcal{L}$ are governed by a BSuMM/OSuMM with look-back(s) $\kappa$, then Algorithm~\ref{alg:fbsearch} asymptotically returns the minimal influencing set.
\end{theorem}

Algorithm~\ref{alg:fbsearch} can accommodate any SuMM as long as summary statistics can be computed from data. 
The way in which it is deployed for BSuMM vs. OSuMM differs in how the summary is specified:
$\mathbf{s}$ is represented by parent configuration $\mathbf{u}$ in BSuMMs and order instantiation $\mathbf{o}$ in OSuMMs. The number of independent parameters $|P|$ can be obtained by multiplying $|\mathcal{X}|-1$ by the summary domain size $|\Sigma_{\mathbf{U}}|$.
\begin{theorem}
\label{thm:complexity}
The 
time complexity of Alg.~\ref{alg:fbsearch} is $O(M^2N)$ where $M$ and $N$ are the number of labels and events, respectively. 
\end{theorem}

\section{Experiments}

The following experiments assess our two proposed SuMMs as well as our learning approach.
Unlike many event sequence datasets in the literature such as books and musical sequences~\citep{rudin2012}, our work is motivated by real-world events and event sequences extracted from text.


\subsection{Influencing Set Discovery}

We conduct an experiment using synthetic data to verify learning capabilities of Algorithm~\ref{alg:fbsearch}. 
A simple event sequence BSuMM dynamics over $5$ event labels is considered,  where the single label of interest has $2$ other labels as its minimal influencing set.
Table~\ref{tab:synth}
displays mean F1 scores, comparing the estimated and ground truth  influencers, over multiple generated sequence datasets as a function of the number of sequences ($K$) generated. The increasing trend shows asymptotic convergence.
Details about the ground truth sequence dynamics and data generation are provided in Appendix~\ref{app:synth_data}.

\begin{table}
\centering
\begin{tabular}{l |c|c|c|c|c}
$K$: & $10$ & $50$ & $100$ & $500$ & $1000$  \\
\toprule 
$F1$: &
$0.23$ & $0.59$ & $0.69$ & $0.93$ & $1$
\end{tabular}
\caption{Mean F1 scores for minimal influencing set discovery in a BSuMM synthetic dataset, as a function of the \# of event sequences ($K$). Monte Carlo error is $\sim 0.003-0.02$.
}
\label{tab:synth}
\end{table}


\subsection{Probabilistic Prediction}

In this experiment, we gauge how the proposed SuMMs compare with some baselines on a (probabilistic) prediction task.

\paragraph{Task \& Metric(s).}

We are concerned with the dynamics of individual labels and therefore conduct an evaluation around probabilistic prediction. 
For any event label $X$, all our models can ascertain the probability of observing the label at any position in the sequence, given the history.
We choose \emph{negative log loss} as the evaluation metric, a popular metric for probabilistic prediction~\citep{bishop2006}.
For our 
binary prediction,
a model's negative log loss is identical to its \emph{log likelihood}. 


\paragraph{Datasets.}

We consider the following structured datasets, some 
derived from time-stamped event datasets where the time stamp is ignored (assumed to be missing or erroneous). 
\begin{itemize}[noitemsep,nolistsep,leftmargin=*]
\item \textbf{Diabetes}~\citep{frank2010}: 
Events for diabetic patients around meal ingestion, exercise, insulin intake and blood glucose level transitions b/w low, medium and high. 
\item \textbf{LinkedIn}~\citep{xu-17-censor}: 
Employment related events such as joining a new role for 1000 LinkedIn users.
\item \textbf{Stack Overflow}~\citep{grant2013}: 
Events for engagement of $1000$ users (chosen from~\cite{du2016recurrent}) around receipt of badges in a question answering website. 
\end{itemize}

Besides these structured datasets, we also experiment with event sequences extracted from (unstructured) textual corpora so as to test on noisy event sequence datasets. Appendix~\ref{app:exps_data} provides further details about curation of these corpora and how they were processed into event sequences:
\begin{itemize}[noitemsep,nolistsep,leftmargin=*]
\item \textbf{Beige Books}: Sequences of topics extracted from 
documents published by the United States Federal Reserve Board on 
events reflecting economic conditions in the United States. 

\item \textbf{Timelines}: 
Sequences of event-related Wikidata~\citep{VrandecicK14} concepts extracted from the timeline sections of event-related Wikipedia articles. 


\end{itemize}

\paragraph{Baselines \& Setup.}
We use two types of similar interpretable parametric models as well as a neural model as baselines:
$k^{th}$ order Markov chains (MC) for varying $k$,
logistic regression (LR) with a varying look-back of $k$ positions for obtaining features, and a simple LSTM \citep{hochreiter1997lstm}.
For experiments, each dataset is split into train/dev/test sets (70/15/15)\%, removing labels that are not common across all three splits. 
Further information about training all models is in Appendix~\ref{app:exps_setup}.

\begin{table*}
\centering
\resizebox{0.75\linewidth}{!}{
\begin{tabular}{l|ccccc|cc|c}
Dataset & 1-MC & 2-MC & 3-MC  & 3-LR & 5-LR  & BSuMM & OSuMM & LSTM \\
\toprule 
Beige Books & -60.91 & -40.37 & -37.66 & -36.85 & -36.15 & \textbf{-36.11} & -38.07 & -63.65  \\
\hline Diabetes & -513.01 & -488.42 & -473.96 & -506.05 & -497.92 & -497.90 & \textbf{-432.89} & -595.57  \\
\hline LinkedIn & -110.58 & -112.55 & -119.55 & \textbf{-92.23} & -93.37 & -114.52 & -115.63 & -135.92  \\
\hline Stack Overflow & -1278.96 & -1283.66 & -1435.12 & -1277.84 & -1263.54 & \textbf{-1242.59} & -1246.64 & -1246.45  \\
\hline Timelines & -154.47 & -611.78 & -1343.52 & -160.84 & -184.05 & \textbf{-141.42} & -142.18 & \textbf{-135.98} \\
\bottomrule
\end{tabular}
}
\caption{Negative log loss (log likelihood),
averaged over labels of interest, for various models computed on the test sets. $k^{th}$ order Markov chains (MC) range from $k=1$ to $3$, 
and logistic regression (LR) is shown for look-back of $k=3$, $k=5$. 
}
\label{tab:results}
\end{table*}

\paragraph{Results.}
Table~\ref{tab:results} shows the negative log loss (identical to log likelihood) on test sets, after averaging over event labels of interest. For the structured datasets and Beigebooks, all labels are of interest; for Timelines, we chose $15$ newsworthy events (ex: `disease outbreak', `military raid', `riot', `war', etc.).
Please see Appendix~\ref{app:exps_baseline_comp} for a more comprehensive comparison with the MC and LR baselines.

BSuMM and OSuMM perform better than the other parametric baselines on $4$ of the $5$ datasets, with BSuMM faring well except on Diabetes.
LR is a strong baseline for the structured datasets because some of these are more suitably modeled by accounting for repetition of previous events. LinkedIn is a prime example with several repeated labels for job role changes in the same company. 
We expected the LSTM to perform better but it may be restricted by the dataset sizes here, which are actually already quite large for the sorts of datasets of interest in this work. Indeed, SuMMs are intended to be applicable to even smaller datasets like those considered in the next subsection, in contrast to neural models which suffer from a lack of interpretability and whose performance generally depends on the amount of training data. LSTM performs best on Timelines, perhaps because this dataset has many more event labels compared to the others and an LSTM might be able to leverage historical information from these more effectively than other models.

The empirical evaluation shows that SuMMs are comparable and often more suitable than the closest baselines, showing flexibility in fitting a variety of event sequence datasets with the additional benefit that they can identify influencers.

\subsection{Case Studies of Influencing Set Identification}

SuMMs are more beneficial for knowledge  discovery  than prediction, since they explicitly indicate influencers of event labels. Investigations on two case studies follow, with details about learning parameters relegated to Appendix~\ref{app:exps_cases}.

\paragraph{Complex Bombing Events.}

We surveyed $15$ Wikipedia articles about planned IED attacks, where each article describes a bombing attempt. 
We manually curated a single sequence of event labels from each article using a small alphabet.
Each sequence is an instance of a `complex event', comprised of simpler primitive events that are represented as event labels in our model. The dataset statistics are as follows:
\# of labels $M = 12$, \# of sequences $K=15$, \# of events $N = 60$.

\begin{figure}
    \centering
    \begin{minipage}{1\columnwidth}
 \centering
 \resizebox{0.7\textwidth}{!}{
 \begin{tabular}{c|c|l}
    Label ($X$) & Influencers ($\mathbf{U}$) & Parameters ($\theta$) \\ 
    \toprule
    denouncement & injury & $\theta_{d|i} = 0.19$ \\ 
    &  & $\theta_{d|\bar{i}} = 0.002$ \\
    \hline
    purchase  & radicalization & $\theta_{p|r} = 0.12$ \\ 
    &  & $\theta_{p|\bar{r}} = 0.002$ \\
    \hline
 sentencing & failure-bomb,   & $\theta_{s|\bar{f},\bar{i}} = 0.06$ \\ 
    & investigation & $\theta_{s|f,\bar{i}} = 0.95$\\
        & &  $\theta_{s|\bar{f},i} = 0.5$ \\
 \bottomrule
 \end{tabular}
        }
 \captionof{table}{Select BSuMM results on small IED dataset.}
   \label{tab:case_study_ied}
    \end{minipage}\hfill
    \begin{minipage}{1\columnwidth}
  \centering \includegraphics[height=1.6in,width=\columnwidth]{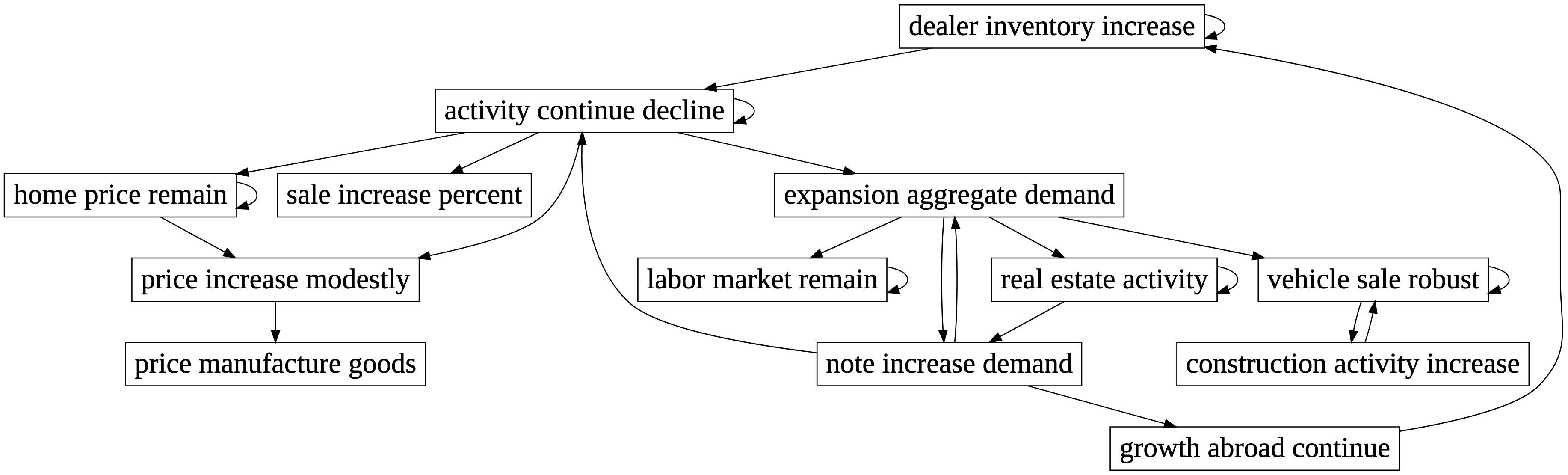}
\caption{Visualization of BSuMM graph over FOMC topics. }
\label{fig:case_study_beigebooks}
    \end{minipage}
\end{figure}

Table~\ref{tab:case_study_ied} 
shows results from learning BSuMM,  indicating selected labels of interest, their influencers and selected parameters.
We use compact parameter notation, e.g. $\theta_{p|\bar{r}}$ denotes the prob. of bombing material `purchase' given that `radicalization' was not observed.
We make a few observations:
\begin{itemize}[noitemsep,nolistsep,leftmargin=*] 
 \item For many labels, we find reasonable single parent labels that make occurrence more likely, for instance, `injury' $\longrightarrow$ `denouncement' and `radicalization' $\longrightarrow$ `purchase'. 
    \item OR  relationships are  discovered, e.g. `sentencing' can occur either when bomb fails to detonate, presumably when the culprit is caught, or when an investigation occurs.  
\end{itemize}



\paragraph{FOMC Economic Conditions.}
We process a subset of the Beige Books dataset mentioned previously. Here we are interested in temporal economic condition trends, therefore we only retain the shorter FOMC statements (which contain fewer topics) and construct a single temporal sequence of topics from all statements, resulting in a dataset with the following statistics:
\# of labels $M = 13$ (out of the 15 original topics), \# of sequences $K=1$, \# of events $N = 676$.

Figure~\ref{fig:case_study_beigebooks} shows a graph learned using 
a BSuMM that is learned for each topic individually, and the learned influencing set is shown as the topic's parents. The graph could help reveal insights about relations, e.g. (i) `activity continue decline' and `expansion aggregate demand' are the most influential economic conditions, with the former inhibiting economic progress -- although the direction of the influence can only be identified from studying model parameters; (ii) some economic conditions are mutually influencing, such as `vehicle sale robust' and `construction activity increase', and `expansion aggregate demand' and `note increase demand' -- in the latter case, the conditions are mutually amplifying. Graphical representations like these from SuMMs could in general be leveraged to assist analysts with knowledge discovery. 


We emphasize that a learned SuMM depends on the chosen event sequence dataset. In these case studies, all influencers are learned in the context of 
small domain-specific  corpora. 
This should always be kept in mind during the analysis so as to deploy the models effectively in practice.

\section{Conclusions}

We have proposed summary Markov models for  dynamics in a sequential event dataset, including two specific models that leverage distinct summary mappings to identify the influence of a set of event labels.  
Experiments on structured datasets as well as event sequences extracted from text illustrate robustness of our models in comparison with prior approaches. 
The main advantage of the proposed models is that they discover influencing sets in addition to  predictive performance comparable with baselines.
The scope of summary Markov models could be expanded by incorporating other approaches to summary mapping such as counting-based models, and adapting parameter sharing ideas from variable order Markov models to allow for expressive models with a reasonable size for the summary range. 
Handling noisy event sequence datasets 
with many irrelevant event labels poses a challenge for future work. 


\section*{Acknowledgments}
We thank the anonymous reviewers for helpful feedback. This research is based upon work supported in part by U.S. DARPA KAIROS Program No. FA8750-19-C-0206. 
The views, opinions and/or findings expressed are those of the authors and should not be interpreted as representing official views or policies of the Department of Defense or the U.S. Government.



\bibliographystyle{named}
{\small
\bibliography{summs}
}

\clearpage

\appendix


\section{Proofs}\label{app:proofs}

\textbf{Proof of  Theorem~\ref{thm:unique}:}

Consider event label set $\mathbf{X}$ and summary function $s(\cdot)$.
We will set up a contradiction for the minimal influencing set. Suppose there are two distinct minimal influencing sets $\mathbf{U}$ and $\mathbf{U'}$.
Then, by definition, for all $s_\mathbf{U} \in \Sigma_\mathbf{U}$, $\tilde{\theta}_{x|h} = \tilde{\theta}_{x|h'}$ for all $h,h'$ consistent with $s_\mathbf{U}$, and for all $s_\mathbf{U'} \in \Sigma_\mathbf{U'}$, $\tilde{\theta}_{x|h} = \tilde{\theta}_{x|h'}$ for all $h,h'$ consistent with $s_\mathbf{U'}$.
Consider the intersection set $\mathbf{U^*} = \mathbf{U} \cap \mathbf{U'}$, with summary mapping range $\Sigma_\mathbf{U^*}$ from all possible histories. Any $s_\mathbf{U^*}$ in $\Sigma_\mathbf{U^*}$ is consistent with a set of $s_\mathbf{U}$ and $s_\mathbf{U'}$. From the equality conditions for the corresponding $s_\mathbf{U}$ and $s_\mathbf{U'}$ sets, $\tilde{\theta}_{x|h} = \tilde{\theta}_{x|h'}$ for all $h,h'$ consistent with this $s_\mathbf{U^*}$. Since this is the case for all $s_\mathbf{U^*}$, $\mathbf{U^*}$ is an influencing set, which results in a contradiction about minimality of the original distinct sets.
Note that the above argument also applies when $\mathbf{U}$ and $\mathbf{U'}$ are non-overlapping sets, in which case the minimal influencing set is actually $\mathbf{U^*} = \emptyset$. Here $\Sigma_\emptyset$ has a single summary state $s_\emptyset$, and $\tilde{\theta}_{x|h} = \tilde{\theta}_{x|h'}$ for all $h,h'$. 

\textbf{Proof of  Theorem~\ref{thm:simplify}:}

Consider the graph $\mathcal{G}^0$ over $\mathcal{L}$ with no edges, which represents that the non-influencing set for each label $X \in \mathcal{L}$ individually is $\mathcal{L}$.
Choose a parameterization $\Theta$ from the $|\mathcal{L}|-1$ simplex, i.e. any set of probabilities $\theta_x$ over all event labels that sum to $1$.
$\Theta$ simplifies according to $\mathcal{G}^0$ by construction. It also simplifies  according to any $\mathcal{G}$ over $\mathcal{L}$, because the parents of each node $X \in \mathcal{G}$ are influencing sets for their children individually, for this $\Theta$; this follows from recognizing that if a label set is an influencing set for label $X$, then any superset is also an influencing set, from the definition of influencing sets.
Note that this result holds for any summary function $s(\cdot)$.

\textbf{Discussion of Remark~\ref{rem:summ_gen}:}

This can be easily verified by setting parent set $\mathbf{U} = \mathcal{L}$ and choosing the appropriate summary domain $\Sigma_{\mathbf{U}}$. In a $k^{th}$ order Markov chain, the domain is of size $|\mathcal{L}|^k$. The choice of parent set would be the same for variable order Markov chains but the domain would be of reduced size, depending on the specific model.

\textbf{Proof of  Theorem~\ref{thm:algo}:}

We provide an outline here but refer the reader to more comprehensive analogous analyses in~\cite{chickering2002optimal} for Bayesian networks and~\cite{gunawardana2016} for graphical event models. For the latter, we refer in particular to Theorem 2 on parametric consistency. 

Suppose the true generating model associated with labels $\mathbf{X}$, i.e. probability for whether the  label in any positions is from this set or not, is from a BSuMM with true minimal influencing set $\mathbf{U^*}$ and known look-backs $\kappa$. Here we also refer to $\mathbf{U^*}$ as minimal parents, analogous to Bayesian networks.
We assume an underlying BSuMM for the BSuMM learner in what follows, but the argument is similar when the underlying dynamics is from an OSuMM and learned using an OSuMM learner.

Parametric consistency, i.e. given enough data and the true parents, parameter estimates from the Bayesian update using summary statistics tend to the true parameters $\theta_{\mathcal{x}|\mathbf{u^*}}$, for all binary instantiations $\mathbf{u^*}$ of $\mathbf{U^*}$.
One can also show structural consistency through a couple of observations.
First, if the current parent set in the algorithm is not
$\mathbf{U^*}$, then adding an event label from $\mathbf{U^*}$  increases the score.
This is because there is improvement in the log likelihood term at the estimated probability parameters (Equation~\ref{eqn:LL}).
This outweighs the constant additional penalty from increasing model complexity in the score in Equation~\ref{eqn:score}.
Furthermore, if the current parent set is a strict superset of $\mathbf{U^*}$, then removing a label that is not in the minimal influencing set increases the score. This is because removing it has no effect on the probability estimates (since it is a non-influencer) and therefore on the log likelihood, but it decreases the complexity component in Equation~\ref{eqn:score}. Thus the greedy search asymptotically returns the minimal influencing set.

\textbf{Proof of  Theorem~\ref{thm:complexity}:}

Each pass in the forward or backward search requires summary statistics and log likelihood computations. If there are currently $P$ parents in the pass, then the summary statistics for either BSuMM or OSuMM can be computed in $O(P N)$, where $N$ is the total \# of events. This is also the time complexity for log likelihood and score computations.
In the worst case, all event labels are added as influencers in the forward pass, with time complexity $\sum_{P=1}^M {O(PN)} = O(M^2N)$, where $M$ is the \# of event labels. The backward pass has the same worst case time complexity.

\section{Experimental Details}\label{app:exps}

\subsection{Sequence Dynamics and Synthetic Data Generation for Influencing Set Discovery}
\label{app:synth_data}

We consider simple BSuMM event sequence dynamics over $5$ event labels, where the label of interest $A$ depends on prior occurrences of labels $B$ and $C$. Label $B$ also depends on prior occurrences of labels $B$ and $C$. $B$ is an amplifier for $A$ whereas $C$ is an amplifier for $B$. Both $A$ and $B$ have look-backs of $3$ positions for BSuMM dynamics. Labels $C$, $D$ and $E$ do not have any historic dependencies and occur independently at any sequence position.

The probabilities are as follows:
\begin{itemize}[noitemsep,nolistsep,leftmargin=*] 
 \item $\theta_{a|\bar{b},\bar{c}} = 0.3$, $\theta_{a|\bar{b},c} = 0.1$, $\theta_{a|b,\bar{c}} = 0.35$,
 $\theta_{a|b,c} = 0.2$. Note that the conditioning in the parameter subscript refers to the binary vector instantiation $\mathbf{u}$ in Definition~\ref{def:bsumm}; for instance,  $\bar{b},c$ is the instantiation where label $B$ does not happen and $C$ happens in their look-back periods.
 \item $\theta_{b|\bar{b},\bar{c}} = 0.1$, $\theta_{b|\bar{b},c} = 0.3$, $\theta_{b|b,\bar{c}} = 0.05$,
 $\theta_{b|b,c} = 0.2$
 \item $\theta_c = 0.3$
 \item $\theta_d = 0.2$
  \item $\theta_e = 0.1$
 \end{itemize}

For synthetic data generation, we generate $K$ sequences (with varying $K$), each of length $10$ events, using the dynamics described above. The BSuMM learning algorithm is deployed to estimate the influencing set of label $A$ from the synthetic data that is generated, and a comparison is made with the ground truth ($\{B, C\}$) to compute the mean $F1$ scores. For the learner, we use the known look-back $\kappa = 3$ positions, prior parameter $\alpha = 0.1$ and penalty weight $\gamma = 1$. Monte Carlo error in the mean $F1$ scores is computed over $10$ runs.

\subsection{Dataset Statistics and Curation Details for Prediction Experiments}
\label{app:exps_data}

Table~\ref{tab:data_summary} summarizes data statistics for all event sequence datasets, both structured and those extracted form textual corpora. Further details about the curation of sequences from the unstructured datasets follow.

\subsubsection{Beige Books}

This collection is curated from 21 years of ``Beige Book" reports\footnote{\url{https://www.federalreserve.gov/monetarypolicy/beige-book-default.htm}} and Federal Open Market Committee (FOMC) statements available as press releases\footnote{\url{https://www.federalreserve.gov/newsevents/pressreleases.htm}}, each published 8 times a year. 
Each sentence in the corpus is mapped to a fixed set of 15 topics in an unsupervised manner, using topic modeling~\citep{Alghamdi2015}. 
In this dataset, a list of extracted topics in a report or statement is considered an event sequence.

\subsubsection{Timelines}

Event-related Wikipedia articles often contain a `timeline' section describing a sequence of more primitive events that have occurred as a part of the `complex event' that is the primary topic of the article. For example, the Wikipedia page on the COVID-19 pandemic in New York City\footnote{\url{https://en.wikipedia.org/wiki/COVID-19_pandemic_in_New_York_City#Timeline}} starts with such a section. It describes the sequences of events around COVID-19 beginning with a description of when and how the first case was observed. We first curate a collection of such timeline sections from Wikipedia articles around significant societal events, e.g. disease outbreaks, protests, and natural disasters. We then map each sentence in each of the timeline sections to a pre-selected set of event-related Wikidata~\citep{VrandecicK14} concepts. This mapping is done through a zero-shot text classification approach similar to the one described in \cite{barker-etal-2021-ibm}. The outcome is a collection of event-related Wikidata concept sequences, where there is a sequence for each timeline.




\begin{table}[t]
\centering
\resizebox{0.7\linewidth}{!}{
\begin{tabular}{l|c|c|c}
Dataset & $M$ & $K$ & $N$  \\ 
\toprule 
 Beige Books & 15 & 349 & 91818 \\ \hline
 Diabetes & $13$ & $65$ & $20210$  \\  \hline
 LinkedIn & $10$ & $1000$ & $2932$  \\  \hline
 Stack Overflow & $22$ & $1000$ & $71254$ \\
 \hline Timelines & $337$ & $529$ & $18549$ \\
\bottomrule
\end{tabular}
}
\caption{Dataset information: \# of event labels ($M$), \# of sequences ($K$) and \# of events ($N$).}
\label{tab:data_summary}
\end{table}

\subsection{Experimental Setup for Prediction Experiments}
\label{app:exps_setup}

Each dataset is split into train/dev/test sets (70/15/15)\%, removing labels that are not common across all three splits. 
The datasets are split randomly based on the sequence identifier, for instance, $70\%$ of patients make up the training set for the Diabetes dataset.
Hyper-parameters for models are tuned using the train/dev combination, and after a model is finalized, evaluation is conducted on the held-out test set.

\subsubsection{MC, SuMMs, LR}

For the MCs, we take a Bayesian approach to estimate parameters, as described earlier for SuMMs, to avoid zero probabilities in the test set. The Dirichlet parameter is selected from $\alpha \in \{ 0.1, 1, 5, 10 \}$. 

For SuMMs, in addition to the $\alpha$ grid as above, we also search over the following hyper-parameter grids: look-back $\kappa \in \{ 1, 5, 10 \}$, assumed to be the same for all labels, and penalty weight $\gamma \in \{0.1, 0.5, 1 \}$.

For LR, LogisticRegression from sklearn was used.

\subsubsection{LSTM}

We create and train the LSTM models using Pytorch.  
Each LSTM consists of three layers: an embedding layer of size 50, an LSTM layer with a hidden dimension of 50, and a fully connected layer. We first transform the data by adding beginning of sequence <BOS>, end of sequence <EOS>, and padding <PAD> tokens to each sequence as needed, to ensure they are all of uniform length and format. These labels do not affect performance as they are ignored in the loss calculation. For each time step in each sequence, we train the model's prediction of the next event label given the sequence up until that point. We normalize the model's output using softmax to obtain a probability distribution for the label of the next event in the sequence across all event labels in the dataset, and then calculate the cross-entropy loss between these predictions and the ground truth. We explore learning rates from 1e-2 to 1e-6 and number of epochs from 10 to 2000 to fine-tune the model hyperparameters  [Table~\ref{tab:lstm_hyperparams}] based on the performance of the dev set during training.

\begin{table}[t]
\centering
\resizebox{0.7\linewidth}{!}{
\begin{tabular}{l|ccccc|cc|c}
Dataset & lr & num epochs \\
\toprule 
Beige Books & 5e-4 & 630 \\
\hline Diabetes & 1e-3 & 400  \\
\hline LinkedIn & 5e-4 & 480  \\
\hline Stack Overflow & 5e-4 & 90 \\
\hline Timelines & 1e-4 & 340 \\

\bottomrule
\end{tabular}
}
\caption{LSTM hyperparameters for each dataset.}
\label{tab:lstm_hyperparams}
\end{table}

\subsection{Learning Parameters for Case Studies}
\label{app:exps_cases}

\subsubsection{Complex Bombing Events}

We learned a BSuMM and OSuMM with hyper-parameters $\alpha=0.1$, $\kappa=10$, $\gamma=0.5$,  chosen to deal with the dataset size; it is desirable to find a balance between model complexity and data availability.
Since the dataset is small, BSuMM and OSuMM learn at most a single parent for most labels and are therefore  identical except for the label `injury'; hence only selected results for the BSuMM are shown in the main text.

\subsubsection{FOMC Economic Conditions}

We learn influencing sets using a BSuMM for each label individually, with $\alpha=0.1$, $\kappa=10$, $\gamma=0.3$. The choice of complexity term $\gamma$ adjusts the sparsity in the graph. A lower value is needed to find influencing sets in smaller datasets.

\begin{table*}[ht]
\centering
\resizebox{0.8\linewidth}{!}{
\begin{tabular}{l|ccccc|cc}
Dataset & 0-MC & 1-MC & 2-MC & 3-MC & 4-MC & BSuMM & OSuMM \\ 
\toprule 
Beige Books & -118.46 & -60.91 & -40.37 & -37.66 & -41.42 &\textbf{-36.11} & -38.07 \\
\hline Diabetes &   -543.05 & -513.01 & -488.42 & -473.96 & -535.66 & -497.90 & \textbf{-432.89} \\
\hline LinkedIn & -133.41 & \textbf{-110.58} & -112.55  & -119.55 & -122.57 & -114.52 & -115.63 \\
\hline Stack Overflow & -1367.27 & -1278.96 & -1283.66 & -1435.12  & -1895.76 & \textbf{-1242.59} & -1246.64 \\
\hline Timelines & -176.31 & -154.47 & -611.78 & -1343.52  & -1656.31 & \textbf{-141.42} & -142.18 \\
\bottomrule
\end{tabular}
}
\caption{Negative log loss (log likelihood), averaged over labels of interest, for various models computed on the test sets. Here the proposed models are compared with $k^{th}$-order Markov chains (MC) for varying $k=0,1,2,3,4$.
}
\label{tab:results_MC}
\end{table*}


\begin{table*}[t]
\centering
\resizebox{0.65\linewidth}{!}{
\begin{tabular}{l|ccc|cc}
Dataset & 3-LR & 5-LR & 10-LR & BSuMM & OSuMM \\
\toprule 
Beige Books & -36.85 & -36.15 & -36.35 &\textbf{-36.11} & -38.07 \\
\hline 
Diabetes & -506.05  & -497.92 & -497.94 & -497.90 & \textbf{-432.89} \\
\hline LinkedIn & \textbf{-92.23} & -93.37 & -93.37 & -114.52 & -115.63 \\
\hline Stack Overflow & -1277.84 & -1263.54 & -1253.04 & \textbf{-1242.59} & -1246.64 \\
\hline Timelines & -160.84 & -184.05 & -200.55 & \textbf{-141.42} & -142.18 \\
\bottomrule
\end{tabular}
}
\caption{Negative log loss (log likelihood), averaged over labels of interest, for various models computed on the test sets. Here the proposed models are compared with logistic regression (LR) for varying look-back $k=3,5,10$. 
}
\label{tab:results_LR}
\end{table*}

\subsection{Detailed Parametric Baseline Comparisons for Prediction Experiments}
\label{app:exps_baseline_comp}

We present a more detailed comparison with the baselines:
$k^{th}$-order Markov chains (MC) (Table~\ref{tab:results_MC})
and logistic regression (LR) (Table~\ref{tab:results_LR}), with more parameter settings than shown in the main text. 

\end{document}